\documentclass[conference]{IEEEtran}
\IEEEoverridecommandlockouts

\usepackage{hyperref}
\usepackage{cite}
\usepackage{amsmath,amssymb,amsfonts}
\usepackage{algorithmic}
\usepackage{graphicx}
\usepackage{colortbl}
\usepackage{float}
\usepackage{caption}
\usepackage{subcaption}
\usepackage{multirow}
\usepackage{multicol}
\usepackage{booktabs}
\usepackage{textcomp}
\usepackage{xcolor}
\def\BibTeX{{\rm B\kern-.05em{\sc i\kern-.025em b}\kern-.08em
    T\kern-.1667em\lower.7ex\hbox{E}\kern-.125emX}}
\begin{document}

\title{Increasing the scalability of graph convolution for FPGA-implemented event-based vision}

\author{
\IEEEauthorblockN{Piotr Wzorek, Kamil Jeziorek and Tomasz Kryjak\thanks{The work presented in this paper was supported by the AGH
University of Krakow project no. 16.16.120.773}}
\IEEEauthorblockA{\textit{Embedded Vision Systems Group, AGH University of Krakow, Poland} \\
\textit{pwzorek@agh.edu.pl, 	kjeziorek@agh.edu.pl, tomasz.kryjak@agh.edu.pl}} 
\and
\IEEEauthorblockN{Andrea Pinna}
\IEEEauthorblockA{\textit{Sorbonne University, France} \\
\textit{andrea.pinna@sorbonne-universite.fr}}
}

\maketitle

\begin{abstract}
Event cameras are becoming increasingly popular as an alternative to traditional frame-based vision sensors, especially in mobile robotics.
Taking full advantage of their high temporal resolution, high dynamic range, low power consumption and sparsity of event data, which only reflects changes in the observed scene, requires both an efficient algorithm and a specialised hardware platform.
A recent trend involves using Graph Convolutional Neural Networks (GCNNs) implemented on a heterogeneous SoC FPGA.
In this paper we focus on optimising hardware modules for graph convolution to allow flexible selection of the FPGA resource (BlockRAM, DSP and LUT) for their implementation.
We propose a ''two-step convolution'' approach that utilises additional BRAM buffers in order to reduce up to 94\% of LUT usage for multiplications. 
This method significantly improves the scalability of GCNNs, enabling the deployment of models with more layers, larger graphs sizes and their application for more dynamic scenarios.

\end{abstract}


\section{Introduction}
\label{sec:intro}


Despite the rapid development of computer vision over the past 60 years many challenges remain -- especially in dynamic environments characteristic for mobile robotics. As an answer to the requirements of vision systems in such applications, the scientific literature increasingly mentions a state-of-the-art neuromorphic sensor inspired by the structure of the human eye -- the event camera or Dynamic Vision Sensor (DVS) \cite{DVS-suvery}.



Unlike traditional cameras which record brightness levels for all pixels at fixed time intervals, DVS records only changes in the observed scene -- independently and asynchronously for each pixel. Each time the change in the logarithm of the brightness of a given pixel reaches a certain threshold, the camera generates an 'event', which is described by four values $\{x, y, t, p\}$,  where $x, y$ are the coordinates of the pixel, $t$ is the timestamp (with a resolution of microseconds), and $p$ is the polarity determining the positive or negative change.


This approach has interesting consequences.
Recording only changes in the scene leads to a significant reduction in redundant information and lower average energy consumption.
Moreover, the high temporal resolution reduces motion blur.
Independent detection of the logarithm of brightness change for each pixel increases dynamic range, enabling effective operation in extreme and rapidly changing lighting conditions.


However, processing event data, which can be described as a sparse point cloud in space-time, in a way that takes full advantage of this sensor is challenging.
To ensure adequate performance and efficiency of event-based vision systems, it is necessary to use appropriate algorithms and hardware platforms.
One potential solution under consideration is the implementation of Graph Convolutional Neural Networks (GCNNs) on heterogeneous FPGAs \cite{jeziorek_graph}.

In this work, we analyse the resource utilisation of graph convolution modules and propose a method to significantly reduce LUT and DSP usage by limiting the number of redundant computations through the use of additional BRAM buffers.
By proposing various module variants that prioritise certain resources (LUT, DSP or BRAM) while minimising the use of others, we enable the implementation of larger GCNNs.



The remainder of this paper is organised as follows. In Section~\ref{sec:related}, we briefly describe the existing work related to the discussed topic.
In Section~\ref{sec:method} we present the methods for using various FPGA resources to implement graph convolution.
In Section~\ref{sec:results} we describe the conducted experiments,  which allowed us to draw the conclusions described in Section~\ref{sec:summary}.

\section{Related Work}
\label{sec:related}

\subsection{Event data processing}


A number of approaches to event data processing have been proposed in the scientific literature. 
The simplest and most common is the generation of ''frame representations'' (event frames), which are analogous to images and are obtained by accumulating events over a specified time window.
Such a frame can then be used at the input of typical computer vision algorithm such as convolutional neural network\cite{DVS-CNN} or recurrent vision transformer \cite{DVS-RT}. 

However, with the growing demand for real-time vision systems in mobile robotics, there is a noticeable trend toward solutions that leverage the spatio-temporal sparsity of event data.
One example are spiking neural networks, which process data in the form of discrete spikes recorded over time \cite{DVS-SNN}. However, the development of this method is currently hindered by challenges in the training process (non-differentiable operations) and still limited access to neuromorphic platforms.
Another option is the use of Graph Convolutional Neural Networks (GCNNs). In these methods, a graph representation is employed, where individual events serve as vertices.
Additionally, edges are defined to connect events that are within a neighborhood determined by a sphere (or hemisphere \cite{DVS-GNN2}) limited by a radius \texttt{R} in three dimensions: x, y and time.
This representation thus preserves information about local dependencies between events.
GCNNs are computationally efficient and their use allows for asynchronous and dynamic updates of the representation for each recorded event \cite{DVS-GNN}.

\subsection{FPGA implementations of GCNNs}
Parallel to algorithms development, methods for accelerating event-based vision for embedded platforms are also being explored. The literature includes works focusing on implementations for embedded GPU platforms (e.g., NVIDIA Jetson), as well as for heterogeneous FPGAs, which, due to their lower energy consumption and flexibility achieved through hardware description languages, offer better performance and real-time task execution \cite{DVS-FPGA}.
In the work \cite{CNN-FPGA}, a data flow for object classification using sparse CNNs was proposed, achieving an accuracy of 72.4\% (N-Caltech dataset) with a latency of 3 ms.
However, CNN-based solutions have high computational complexity, which leads to significant FPGA resource consumption, making implementations for embedded systems challenging.
To address this issue, researchers explore hardware implementations that leverage the sparse nature of the data, including those that utilise GCNNs.



In \cite{GNN-Holandia}, a graph generation and convolution method for processing event-based data on FPGA is proposed.
The system achieves high accuracy for one-class classification (87.8\% for N-Cars) and a low average latency of 16 $\mu$s (variable due to the use of external memory resources).
However, this system does not support MaxPool layers, which are essential for larger models and more complex computer vision problems.

Support for FPGA implementation of all layers characteristic of GCNN is provided in \cite{jeziorek_graph}.
The system does not use external memory resources for inference, ensuring constant and deterministic latency (4.47 ms) and high throughput (13.3 million events per second).
The authors have made the code available in an open-source repository.
In the next sections, we describe a proposal for modifying the graph convolution layers used in this work to address the limitations related to FPGA resource utilisation for larger models.

\section{The proposed method}
\label{sec:method}

\begin{table*}[]
\centering
\caption{The number of necessary parallel multipliers for the Baseline and Two-Step methods, depending on \texttt{TIME\_WINDOW}, the \texttt{SIZE} of the graph after MaxPool and the dimension of the output feature map (\texttt{OUT}), along with the resulting reduction.}
\resizebox{0.98\textwidth}{!}{%
\begin{tabular}{@{}ccc|ccc|ccc|ccc@{}}
\toprule
\multirow{2}{*}{Time Window} & \multirow{2}{*}{Size} & \multirow{2}{*}{Throughput} & \multicolumn{3}{c}{Parallel Multiplications (baseline)} & \multicolumn{3}{c}{Parallel Multiplications (two-step)} & \multicolumn{3}{c}{Decrease [\%]} \\ 
&&& OUT 32 & OUT 64 & OUT 128 & OUT 32 & OUT 64 & OUT 128 & OUT 32 & OUT 64 & OUT 128 \\ \midrule
50 ms & 32 & 312500 & 2 & 4 & 8 & 1 & 1 & 1 & 50 & 75 & 88   \\ 
50 ms & 64 & 156250 & 16 & 32 & 64 & 1 & 2 & 4 & 94 & 94 & 94    \\ 
100 ms & 32 & 625000 & 2 & 2 & 4 & 1 & 1 & 1 & 50 & 50 & 75      \\ 
100 ms & 64 & 312500 & 8 & 16 & 32 & 1 & 1 & 2 & 88 & 94 & 94      \\ \midrule
100 ms & 128 & 156250 & 64 & 128 & 256 & 8 & 16 & 32 & 88 & 88 & 88     \\
30 ms & 32 & 187500 & 4 & 8 & 16 & 1 & 1 & 1 & 75 & 88 & 94     \\ 
30 ms & 64 & 93750 & 32 & 64 & 128 & 2 & 4 & 8 & 94 & 94 & 94     \\ \bottomrule
\end{tabular}%
\label{table:theory}
}
\end{table*}

\subsection{Overview}

The system described in \cite{jeziorek_graph} uses a directed 3D graph of fixed size \texttt{INPUT\_SIZE} (either 128×128×128 or 256×256×256), which represents a fixed \texttt{TIME\_WINDOW} of data (100 ms or 50 ms) for classification.
However, this representation is updated continuously and the data is processed asynchronously in the first part of the system -- event by event.


The recorded events (that appear at the input at intervals consistent with their timestamps) are first transformed into vertices by scaling their $x$, $y$ and $t$ values accordingly. They are then connected by edges to previously captured vertices located at a distance of no more than \texttt{R=3} in each of the three dimensions ($x$, $y$ and $t$).



The vertices, along with the list of edges, are then passed to the convolution module. The system uses the \texttt{PointNetConv} convolution, which can be described by the following formula:
\begin{equation}
\hat{x_i} = \left (\underset{j \in N(i)}{\max} \phi (x_j, p_j - p_i)\right )
\end{equation}
For each incoming vertex, the result of the linear layer $\phi$  is computed for its feature map $x_j$ (the so-called self-loop), as well as for the feature map of each vertex in the neighborhood $N(i)$ connected to it by an edge. For the first convolution, the feature map is equal to the polarity of the event. Each feature vector is augmented by the difference in positions of the connected vertices $(p_j - p_i)$ in three dimensions (for self-loop it is $(0, 0, 0)$). The output feature map is determined by applying the element-wise $max()$ operation to all $\phi$ linear layer results for each vertex.


The next layer is the so-called ''Relaxing MaxPool 4×4'' where graph representation is scaled down to extract a wider context of the data and to reduce the necessary memory resources for storing feature maps (\texttt{SIZE} = \texttt{INPUT\_SIZE/4}).
Additionally, the feature vectors (along with the list of edges) are organised into so-called ''temporal channels'' (\texttt{TC}) -- arrays implemented in BRAMs of size \texttt{SIZE}×\texttt{SIZE}, which represent data recorded during specific subsets of the time window (\texttt{TIME\_WINDOW}/\texttt{SIZE}).
Data is transferred to the subsequent (synchronous) part of the system only after all events corresponding to that subset time window are processed.



Based on the network's hyperparameters, including the time window and graph size after MaxPooling, we can determine the precise time interval between successive \texttt{TC}.
This allows us to calculate the necessary throughput for each subsequent graph convolution layer.
The main strategy for reducing FPGA resource consumption in the system is to perform part of the calculations sequentially, without affecting the latency of the solution.
The number of possible sequential multiplications depends on the required throughput for the particular layer.

\subsection{The baseline convolution}

The graph convolution module in the sequential part of the system proposed in \cite{jeziorek_graph} operates as follows: after scaling the graph (MaxPool 4×4), the neighbourhood radius for edges is also scaled (in the sequential part, \texttt{R=1}).
As a result, each vertex has a maximum of 17 edges (8 candidates in the currently processed \texttt{TC} and 9 candidates in the previous one).
The input feature vectors are read from two BlockRAM modules that store both \texttt{TC}s containing candidates. 
The module reads 18 values for each vertex.
Using two memory modules the system processes two elements in parallel.

Calculating a single element of the output feature map involves expanding the input feature vector by the position difference ($\Delta x$, $\Delta y$, $\Delta t$), multiplying it by the corresponding column of the weight matrix (with LUT resources), and then requantising the resulting value (with DSP multipliers).
To determine the entire output feature vector, this step must be repeated for each of its elements.
In summary, by employing two parallel modules for vector multiplication, the number of clock cycles $N_{CC}$ required to perform the calculation is:

\begin{equation}
N_{CC} = SIZE \times SIZE \times 9 \times OUT\_DIM
\end{equation}

where: \texttt{SIZE} is the size of the currently processed graph, and \texttt{OUT\_DIM} is the number of elements of the output feature vector.
At the same time, the required throughput of a layer is determined by the frequency of new \texttt{TC} generation at the input, expressed by the following formula:

\begin{equation}
T_{CC} = \frac{TIME\_WINDOW\_NS}{SIZE \cdot NS\_PER\_CLK} 
\end{equation}

where: $NS\_PER\_CLK$ is the period of a single clock cycle expressed in nanoseconds (in this work 5 ns -- clock frequency of 200 MHz).
In cases where $N_{CC} > T_{CC}$, it is not possible to compute each element of the output feature map separately and it becomes necessary to increase the number of parallel vector multiplication modules.
The number of parallel multipliers needed, depending on the value of the time window and the size of the graph, is shown in Table \ref{table:theory}.
We have considered the layers used in the work \cite{jeziorek_graph} (\texttt{TIME\_WINDOW} 50/100 ms, \texttt{SIZE} 32/64).
Additionally, anticipating future developments of the described system, its capabilities were evaluated also for data with higher resolution (\texttt{SIZE}=128) or higher dynamics of the observed scene (\texttt{TIME\_WINDOW}=30ms).

\subsection{Utilisation of DSP}
The vector multiplication in the graph convolution module is implemented using LUT resources, which consequently causes a significant increase in the consumption of logic resources when the described method is used for smaller time windows and larger input graphs.
The simplest method to reduce LUT consumption, as pointed out in \cite{jeziorek_graph}, is to implement some of the multiplications using DSP modules. When the logic resource consumption approaches 100\%, the use of a vector multiplication module utilising DSPs instead of the baseline can enable the implementation of larger GCNN models.
It should be noted, however, that DSP resources are limited for embedded SoC FPGAs (e.g., 1728 for the ZCU104 board with XCZU7EV-2FFVC1156 chip).
Therefore, we propose an additional strategy aimed at reducing the utilisation of both LUTs and DSPs by leveraging a third resource -- BlockRAM.

\subsection{Two-step graph convolution}

\begin{figure}
    \centering
    \includegraphics[width=0.35\textwidth]{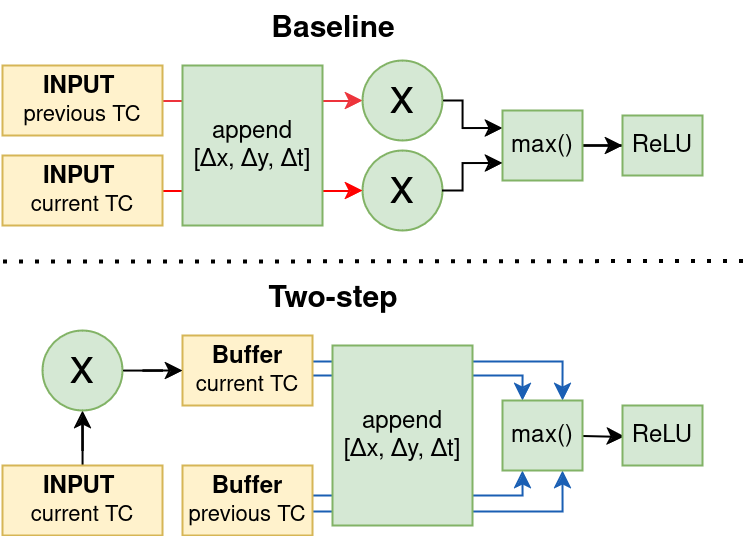}
    \caption{The baseline and two-step methods. Yellow blocks represent BRAM memory, and green ones represent logical operations. X represents both multiplication and requantisation. The colour of the arrows indicates the number of memory reads for a single vertex -- red is 9, black is 1, and blue is 5.}
    \label{fig:diagram}
\end{figure}

Based on the analysis of the graph convolution module from \cite{jeziorek_graph}, we have determined that the same input feature maps are processed by the vector multiplication modules multiple times — both when they are processed individually (self-loops) and when a vertex connected to them by an edge is processed.
This redundancy limits the number of possible sequential multiplications.
To address this issue, we propose a two-step graph convolution module.
In this method, the first step involves performing multiplications for all elements of the processed \texttt{TC}.
At this stage, the feature vectors are not extended with the position differences ($\Delta x, \Delta y, \Delta t$) -- only self-loops are computed.
The output feature maps are stored in two buffers implemented using BRAM memory: one for the currently processed \texttt{TC} and one for the previous one.

After completing the first step, the module proceeds to the second one, in which the feature maps corresponding to each vertex and its connected neighbours are read from the buffers.
Due to the use of two dual-port buffers, reading all 18 elements (the vertex and its 17 neighbours) requires only 5 clock cycles.
Since in the first step, the input vectors were not extended with the position differences ($\Delta x, \Delta y, \Delta t$), before the $max()$ operation, it is necessary to add to each feature vector the result of multiplying these three values by their corresponding weights.
Since these results are constant for a given layer (the values of $\Delta x$, $\Delta y$, and $\Delta t$ can only take values ${-1, 0, 1}$), they can be stored in a Look-Up Table.
In summary, by applying an additional buffer and splitting the graph convolution into two stages, the number of required clock cycles (with a single vector multiplication module) is reduced to:
\begin{equation}
N_{CC} = SIZE \times SIZE \times (OUT\_DIM+5)
\end{equation}


In cases where $T_{CC} < N_{CC}$, it is still necessary to use more than one multiplier.
Table \ref{table:theory} presents the number of required parallel multipliers for different layers and models. 
Its worth noting that their number is always at least twice smaller than that of the baseline method. 
For architectures where LUT utilisation approaches 100\% and BRAM modules are available the two-step convolution method facilitates their implementation.
Figure \ref{fig:diagram} visualises the differences in the operation of the baseline and two-step methods.

\section{Experimental results}
\label{sec:results}

In order to confirm the validity of the proposed modifications to the model, we carried out tests in both the software model and the hardware implementation. 

The aim of the software experiments was to verify the impact of the two-step method on the performance of the classification system.
It should be noted that in the baseline method, the requantisation is performed after the entire multiplication is completed.
In the two-step method, on the other hand, the feature map elements are requantised before adding the difference-in-position factor.
The rounding applied can lead to variations in the results, specifically affecting the least significant bit ($\pm1$ for an 8-bit value).
Software experiments (involving the processing of the N-Caltech dataset) confirmed that such differences appear, on average, in 14.3\% of the output feature map elements of a given layer (with a standard deviation of 0.035).
However, the evaluation of model accuracy after modifications showed that even replacing more than one baseline layer with a two-step layer leads to a change in accuracy of no more than $\pm0.3\%$.

To investigate actual resource utilisiation, we prepared a simple GCNN consisting of graph generation (\texttt{TIME\_WINDOW=50ms}, \texttt{INPUT\_SIZE=256}), asynchronous convolution $4 \rightarrow 16$, a MaxPool 4x4, and synchronous convolution in two variants: $16 \rightarrow 32$ and $16 \rightarrow 64$.
Each of the evaluated methods was tested with Vivado simulation for consistency with the software model and then implemented for the ZCU104 board with a 200MHz clock (without any timing issues).
The resource utilisation after implementation is presented in Table \ref{table:practice}.
Each method leads to an increase in the usage of one resource with a simultaneous reduction of another.
The application of the two-step method resulted in a reduction of LUT resources in the GCNN implementation by 70\% (for convolution $16 \rightarrow 32$) and 78\% (for $16 \rightarrow 64$), respectively.
The DSP-conv method, on the other hand, allowed for a reduction of 64\% and 71\%, respectively.

\section{Summary}
\label{sec:summary}

\begin{table}[!t]
\centering
\caption{GCNN's resource utilisation depending on applied method with the most heavily utilised resource in bold.}
\resizebox{0.48\textwidth}{!}{%
\begin{tabular}{@{}c|ccc|ccc@{}}
\toprule
\multirow{3}{*}{Method} & \multicolumn{6}{c}{Resource utilisation (ZCU104)} \\
& \multicolumn{3}{c}{$16 \rightarrow 32$ convolution} & \multicolumn{3}{c}{$16 \rightarrow 64$ convolution} \\
& LUT & BRAM & DSP& LUT & BRAM & DSP \\
\midrule
Baseline & \textbf{36425} & 73 & 96 & \textbf{65634} & 73 & 160 \\
DSP-conv & 13080 & 73 & \textbf{416} & 18761 & 73 & \textbf{768} \\
Two-step & 10776 & \textbf{117} & 83 & 13966 & \textbf{174} & 90 \\
\bottomrule
\end{tabular}%
\label{table:practice}
}
\end{table}
In this work, we have proposed a ''two-step graph convolution'' method applicable to a GCNN implemented on an FPGA for event data processing.
It allows to reduce the LUT usage for multiplication by 50-94\% (depending on the configuration) by using additional BRAM buffers.
Furthermore, the approach has a negligible impact on classification accuracy of no more than $\pm$0.3\% (for the N-Caltech dataset).
Based on an analysis of resource availability in the target device, appropriate variants of each graph convolution module, that prioritise a specific resource -- LUT, BRAM, or DSP, reducing the usage of others, can be selected, allowing the implementation of larger networks for larger graphs and smaller time windows.
We plan future work on GCNN implementation in FPGAs for more challenging datasets and more complex computer vision tasks (e.g. object detection).



\end{document}